# Cascade Neural Ensemble for Identifying Scientifically Sound Articles

Ashwin Karthik Ambalavanan, Murthy Devarakonda
Arizona State University

## Abstract

Background: A significant barrier to conducting systematic reviews and meta-analysis is efficiently finding scientifically sound relevant articles. Typically, less than 1% of articles match this requirement which leads to a highly imbalanced task. Although feature-engineered and early neural networks models were studied for this task, there is an opportunity to improve the results.

Methods: We framed the problem of filtering articles as a classification task, and trained and tested several ensemble architectures of SciBERT, a variant of BERT pre-trained on scientific articles, on a manually annotated dataset of about 50K articles from MEDLINE. Since scientifically sound articles are identified through a multi-step process we proposed a novel cascade ensemble analogous to the selection process. We compared the performance of the cascade ensemble with a single integrated model and other types of ensembles as well as with results from previous studies.

Results: The cascade ensemble architecture achieved 0.7505 F measure, an impressive 49.1% error rate reduction, compared to a CNN model that was previously proposed and evaluated on a selected subset of the 50K articles. On the full dataset, the cascade ensemble achieved 0.7639 F measure, resulting in an error rate reduction of 19.7% compared to the best performance reported in a previous study that used the full dataset.

Conclusion: Pre-trained contextual encoder neural networks (e.g. SciBERT) perform better than the models studied previously and manually created search filters in filtering for scientifically sound relevant articles. The superior performance achieved by the cascade ensemble is a significant result that generalizes beyond this task and the dataset, and is analogous to query optimization in IR and databases.

## Introduction

MEDLINE and EMBASE are large repositories of scientific articles that are often searched for systematic reviews and other meta-analysis. MEDLINE has about 27 million articles at this time and a million new articles are being added every year. However, less than 1% of these studies meet the criteria for relevant scientifically sound articles (see Table 1 for an example of such criteria).[1] Finding very small number of relevant studies from massive repositories is a significant burden in effectively unlocking the existing knowledge in such repositories.

The standard approach for finding relevant articles in MEDLINE was to use *Clinical Query filters* which were rules based on combinations of text strings, MeSH (Medical Sub Headings), and database tags.[2,3] The filters were optimized by expert librarians using statistical analysis of term counts and tags in relevant and irrelevant articles. Although the filters were updated over time, the accuracy of the results had not changed significantly over two decades.[4] Some useful database tags (e.g. MeSH terms) become available only after significant time lag (e.g. 17 to 328 days) after the articles were added.[5]

Early studies of machine learning approaches[6–8] for this task have used extensive feature engineering, including proprietary and time-dependent features. Two recent studies[5,9] have shown that the early neural networks models (CNN) can outperform manually created search filters without feature engineering. However, recently developed pre-trained context-based neural language models, BERT (Bidirectional Encoder Representations using Transformers)[10] and its variants,[11–15] have not been studied for this task. These new neural models have established new state-of-the-art results for several general domain tasks[10] and for biomedical information extraction tasks.

Furthermore, since the logic for finding relevant scientifically sound articles involves multiple distinct constraints (see Table 1), an interesting research question is how to model the constraints. Is a single model or an ensemble of constraint models the most optimal? While ensembles have been frequently used before, they were often used to model the same task using different methods and then aggregate their predictions. Here we study different techniques for ensembling constraint models. Our results generalize to effective ways of modeling multiple constraints of any biomedical NLP task.

Information retrieval (IR) challenges in the biomedical domain have been well documented.[16–19] In recent years, neural networks have also been used for information retrieval.[20–23] However, the task here can be better modeled as a classification problem rather than as an IR problem since ranking the results was not needed.

*Table 1. Manual annotations for each article in Clinical Hedges.*

| Criterion Name | Description |
|---|---|
| Format | Whether it is an original study, review, case report, general and miscellaneous article. |
| HHC | Whether the article is of interest to human health care |
| Purpose | Whether the article is about etiology, prognosis, diagnosis, treatment (or prevention), costs, economics, disease related prediction, qualitative study, or something else |
| Rigor | Whether the study meets experiment design quality criteria specific to a purpose, e.g. random allocation to comparison groups if purpose is treatment. |

We used the dataset from the Clinical Hedges project,[24] which pioneered work in this area by extracting a dataset of about 49,000 articles from MEDLINE and manually annotating each article on four different criteria that were shown in Table 1. The criteria can be combined to identify scientifically sound articles meeting flexible requirements, for example, scientifically sound articles describing diagnosis methods (rather than treatment methods) in human healthcare.

Our study showed that a novel cascading ensemble of SciBERT models outperformed the other architectures and our results set a new state-of-the-art performance standard for this task on the Clinical Hedges dataset. The cascade ensemble achieved 0.7505 F measure on a previously-studied[5] subset of the Clinical Hedges dataset, and 0.7696 F measure on the full dataset.

While caution should be used in comparing the results because of certain experimental differences (discussed in the paper), the cascade ensemble achieved 0.2405 F measure improvement in absolute terms (on the 1.00 scale) and an impressive 49.1% error rate reduction compared to the CNN model (F measure 0.510) reported in Del Fiol et al.[5] The improvement relative to the study in Marshall et al,[9] where the full dataset was used, was 0.0581 F measure in the absolute terms (on the 1.00 scale) and 19.7% error rate reduction relative to the best performing model (CNN + Publication Tag ensemble, F measure 0.7058) of the study.

## Methods

### Dataset

The Clinical Hedges dataset consists of 49,028 unique articles retrieved from MEDLINE in 2000 (and 50,590 annotations).[24] Each article was identified by its MEDLINE id, called PMCID, which allowed us to fetch the article data, title and abstract, as well as the other metadata from MEDLINE. Each article was manually rated on four criteria: (1) The "Format" of the article, which can be original study, review, case report, or general and miscellaneous studies; (2) Whether it is concerning human healthcare or not, abbreviated as "HHC"; (3) Intended "Purpose" of the paper such as treatment, diagnosis, prognosis, and economics; (4) Whether the study has scientific "Rigor" in that if it met certain study design constraints that are specific to the purpose, i.e. if the purpose is treatment then the study must allocate subjects randomly to study groups.

The annotators were instructed to rate each article on a sequence of criteria starting from criterion 1 to criterion 4. They were asked to stop rating an article further based on the rating outcome at a criterion. For example, when the Format of an article was determined as "general and miscellaneous" type, raters were asked not to rate it further.

Because the articles were rated on multiple criteria, the number of positive and negative articles depended on the specific combination of criteria specified for a task. If the task was to identify human healthcare related original studies of treatment conducted with rigor, then the positive class contained 1,587 articles. The negative samples were therefore, 49,003, resulting in a highly imbalanced dataset. Other selection criteria such as diagnosis or prognosis as the purpose was possible and would result in a different number of positive and negative samples.

Publication type is a metadata tag that was manually assigned to articles in MEDLINE. Although these tags could potentially give an insight into the relevance of the article, not all articles were assigned this tag. They were slow to be generated (can take up to 6 months for tags to be put on an article after publishing)[9] and our analysis showed that useful tags were present for about 2000 articles of the Clinical Hedges dataset. Therefore, we could use it only as a noisy input feature (i.e. cannot always be relied on) for an article.

## Models

We used uncased SciBERT, which is an uncased BERT model pre-trained on a corpus of scientific articles, as the core model in our study. The model was pre-trained on a random sample of 1.14M papers from Semantic Scholar (semanticscholar.org). The pre-training corpus, therefore, closely resembled MEDLINE articles, and consisted of full text of papers, 18% from the computer science domain and 82% from the broad biomedical domain. The resulting corpus had 3.17B tokens, about the same as the 3.3B BERT pre-training tokens, but had only 42% words in common with it. Since there were four different criteria for classifying an article as a scientifically sound study, we conceived four different ways the task can be modeled.

### Individual Task Learner (ITL)

This is the basic form of the neural network model where a single pre-trained SciBERT model was used with a feed forward network (FFN) as the text classification head (see Figure 1). The logits from the CLS tag of SciBERT are input to the FFN. The integrated model predicts whether the article input to it meets all the criteria for the positive class or not. Thus, a single model learns the integrated criteria for classifying an article.

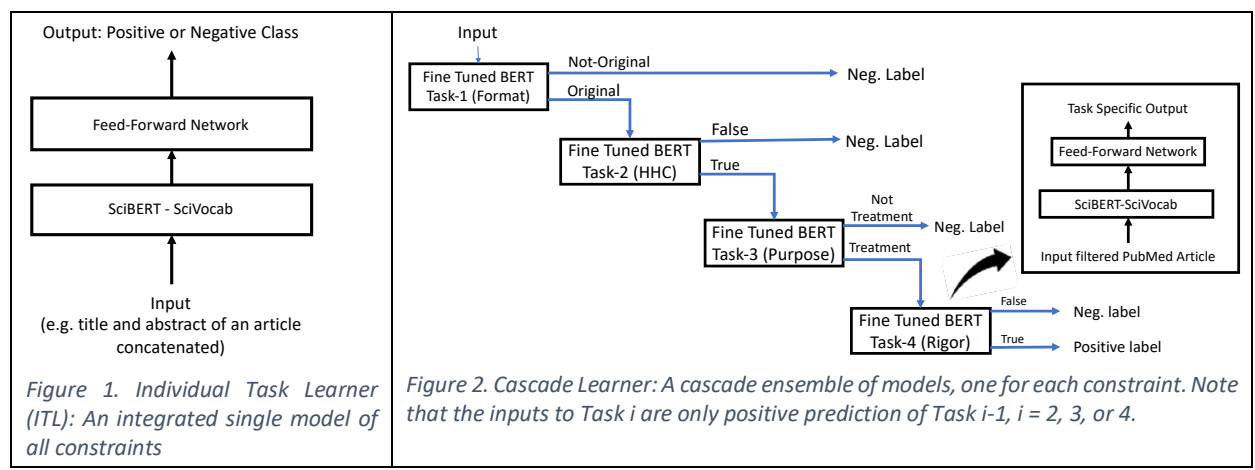

*Figure 1. Individual Task Learner (ITL): An integrated single model of all constraints*

*Figure 2. Cascade Learner: A cascade ensemble of models, one for each constraint. Note that the inputs to Task i are only positive prediction of Task i-1, i = 2, 3, or 4.*

### Cascade Learner

In this architecture, we train four different SciBERT models each having the same architecture as the ITL described above but each learning only one of the four criteria (see Figure 2). The first model (Task-1) is trained on classifying an article based on the Format attribute alone. Similarly, the second (Task-2) was trained to classify on HHC, the third (Task-3) on Purpose, and the fourth (Task-4) on Rigor. The (sub)models are ensembled as a series of cascading blocks like a water fall. A model only sees the articles that were assigned a positive label by the previous model in the cascade. The articles that were classified as negative by the previous model are not further analyzed, and were assigned a negative label. The first model sees all the articles in the dataset.

For example, if an article was classified by Task-1 model as "Not Original", then the sample is not analyzed further and the label for it is finalized as negative. On the other hand, if an article was classified as positive by the Task-1 model, it is further analyzed by Task-2 (and the subsequent models depending on the outcome from Task-2). This was done during the training as well as in the test. So, downstream models are trained and tested on only a subset of the data. Each component of the cascade ensemble was trained separately on inputs and labels corresponding to it. The intuition behind this approach is twofold. First, this is how the manual annotators labeled the data. Second, each model learns to predict a particular criterion well under the narrow conditions of its filtered input.

### Traditional Boolean Ensemble Learner (Ensemble-Boolean)

Ensembling multiple models is a well-known technique where multiple models independently analyze each input and the final prediction is made by combining the output of all models. In fact, the Cascading Learner is a novel variation of ensembled models. In this architecture, each of the 4 models were trained and tested on the entire dataset rather than on a subset. We combine the outputs using Boolean logic (i.e. a conjunction of all outputs) according to the criteria for selecting scientifically sound articles (see Figure 3). So, the final output label is False if any of the models classifies it as False.

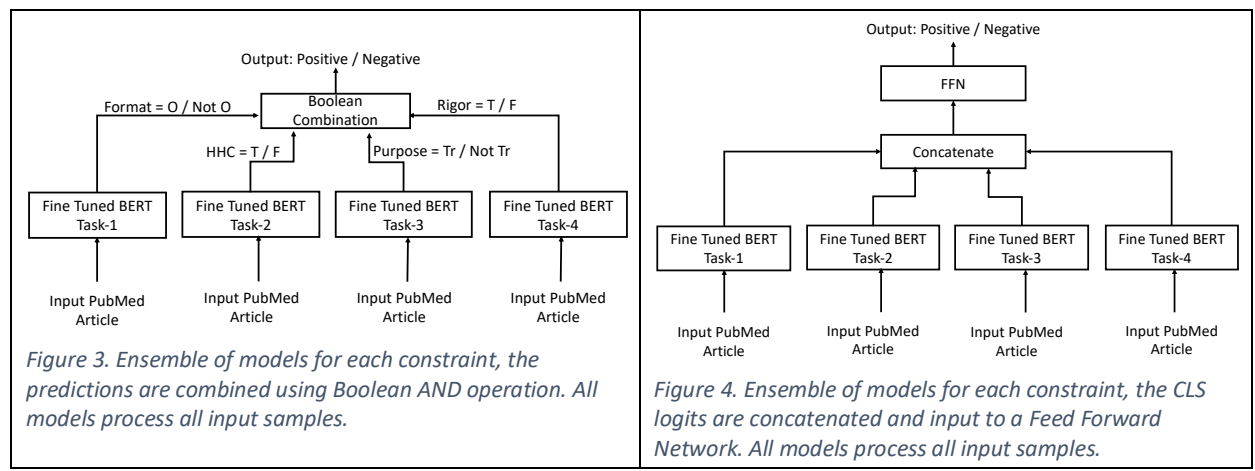

*Figure 3. Ensemble of models for each constraint, the predictions are combined using Boolean AND operation. All models process all input samples.*

*Figure 4. Ensemble of models for each constraint, the CLS logits are concatenated and input to a Feed Forward Network. All models process all input samples.*

### Feed-Forward Network Ensemble Learner (Ensemble-FFN)

Unlike the previous architecture, the intuition for this architecture is to allow the ensemble to learn how to combine individual predictions with the help of an FFN to make the final prediction. We fine-tuned a SciBERT model, one for each of the 4-tasks, by training on the tasks separately. The CLS embeddings obtained from each of these trained sub-models were concatenated together (forming a 4x768 vector) as the input to the feed forward network. The FFN then decides whether to give a positive or negative label to the input article based on the concatenated embeddings (see Figure 4). The entire model was trained together allowing error propagation through the FFN and all the individual models at the same time.

### Experiments

The goal of our experiments is to study how different ensembles and ITL performed on the task of classifying scientific articles based on the four filtering criteria as used in the Clinical Hedges dataset. Since the number of positive samples is small, we used 10-fold cross validation to make the best use of the samples. As is conventional, we randomly split the positive and negative samples into 10 folds, tested the model on the i[th] fold after training the ensemble on the remaining 9 folds, and varied i from 1 to 10. The test results were micro averaged across all the folds to calculate the model precision, recall, and F measure.

The input to a model is a concatenation of the title and abstract of an article, in that order. The title and abstract lengths varied depending on the article, but SciBERT has a maximum input sequence length of 512. So, we conducted experiments with different maximum sequence lengths. Similarly, we also experimented with different ratios of positive to negative samples. Lastly, we studied the effect of including the Publication Type tag in the input by concatenating the tag text to the input at the front of the sequence.

*Data subsets:* We experimented with two subsets of the Clinical Hedges dataset that were used in two most recent studies - Del Fiol et al[5] and Marshall et al[9] – for comparison purposes. Marshall et al used all the samples (full set) from the dataset for testing, but Del Fiol et al used a subset that met the following constraint:

*Format = Original / Review / Blank, HHC = True / False, Purpose = Any, Rigor = True / False*

As a result, the Del Fiol set had only 28,331 articles whereas the Marshall set had all 50,590 articles (see Table 2). Both studies used the following identical definition for positive samples:

*Format = Original, HHC = True, Purpose = Treatment and Rigor = True*

Furthermore, the two studies used different training data, i.e. different selections of MEDLINE articles (extracted using PubMed search with a specific query) as noisy training data. We refer the reader to the publications for details, but briefly, articles returned by PubMed for the query term containing "Randomized Controlled Trials" in a particular time period were used as positive samples and all the remaining articles in the same time period were used as negative samples for training.

*Table 2. Dataset statistics of the two subsets we used.*

| Data subsets used | Positive Samples | Negative Samples | Total Samples |
|---|---|---|---|
| Del Fiol et al | 1,553 | 26,778 | 28,331 |
| Marshall et al | 1,587 | 49,003 | 50,590* |

*Experiments with ensembling techniques:* We conducted the experiments with both subsets of the Clinical Hedges dataset. We present results separately for each subset and compare changes in performance as the number of samples increased. It should be re-iterated that unlike the Del Fiol et al and Marshall et al studies, we used the 10-fold cross validation (on each set separately) rather than using noisy training data, which allowed robust training and evaluation on a highly imbalanced dataset. In all cases, we present results only for the positive class, which is the only class of interest. We also studied performance on individual criteria for the ensembling to quantify which criterion are most challenging to accurately model.

*Sampling ratio experiments:* The ratio of positive and negative samples is 1:18 in the Del Fiol et al subset of the Clinical Hedges dataset and it's almost 1:32 in the full set. So, we conducted experiments to study the impact of various sampling sizes and ratios in training the models. (Testing was done on all samples in each of the datasets.) As needed, we randomly down-sampled negative samples and up-sampled positive samples through duplication in the experiments. It should be noted that as the number of samples increased, computing resources needed for training and testing the ensembles increased unacceptably high (multiple days of running time and GPU out of memory problems), and so we studied the impact of sampling ratios on ITL only. Furthermore, because of its architecture, changing sampling ratios in Cascade Learner makes comparison with other models and interpretation difficult and so we always used balanced 1:1 positive and negative sample ratios at all components of the ensemble in training the model.

Sequence length experiments: We observed that the titles and abstract vary significantly in length. As shown in Table 3, while the average length is 178.82 words, the maximum length is 856 words. The 95th, 92nd, and 69th percentiles were 420, 384, and 256 respectively. Note that the SciBERT standard (base) model allows sequence lengths up to 512 and but in general increasing sequence length increases the computational (GPU memory and running time) requirements. So, we used the default sequence length of 256 in the initial configuration and studied the performance impact of increasing the sequence length to 384.

Table 3. Sequence length statistics for title + abstract for articles in Clinical Hedges

| Statistic | Sequence Length (in words) |
|---|---|
| Average Length | 178.82 |
| 69th Percentile | 256 |
| 92nd Percentile | 384 |
| 95th Percentile | 420 |
| Maximum Length | 856 |

Publication Type (PT) tag experiments: As mentioned earlier, some MEDLINE articles are manually assigned one or more publication type tags (https://www.nlm.nih.gov/mesh/pubtypes.html) sometime after (typically several months) the article was added to the repository. We used the publication tags on an article as a part of the input text, i.e. we prepended the tags to title and abstract text.

## Results

Performance comparison of the ensembles and ITL: The first set of results (see Table 4) show performance comparison of the ensemble architectures and ITL. The Table presents precision, recall, and F measure for the four architectures, separately for the two subsets. Here we used sequence length of 256 words, randomly down-sampled negative samples to the number of positive samples, and PT tag was not used. The Cascading Learner outperformed the rest of the models for both datasets. Especially, it achieved the highest precision (0.6551 and 0.6600) and the highest F measure (0.7505 and 0.7542) for each dataset. The ITL has outperformed all models in terms of recall, achieving 0.9627 and 0.9616 for the datasets respectively.

Table 4. Performance of the ensembles and ITL on the Clinical Hedges dataset

| Model | Partial Dataset | | | Full Dataset | | |
|---|---|---|---|---|---|---|
| | Precision | Recall | F Measure | Precision | Recall | F Measure |
| ITL | 0.5137 | **0.9627** | 0.6700 | 0.3825 | **0.9616** | 0.5472 |
| Cascading Learner | **0.6551** | 0.8783 | **0.7505** | **0.6600** | 0.8796 | **0.7542** |
| Ensemble-Boolean | 0.5608 | 0.8847 | 0.6865 | 0.4728 | 0.9200 | 0.6246 |
| Ensemble-FFN | 0.4206 | 0.9446 | 0.5820 | 0.3177 | 0.9452 | 0.4756 |

One other interesting observation is that as the number of negative samples increased in test, as we go from the partial dataset to the full dataset (recall that the difference between them is only negative samples considered from Clinical Hedges), performance of all models reduced, except Cascade Learner. In fact, Cascade Learner performance improved slightly. Clearly, filtering out negative samples early using simpler constraints works more effectively.

Performance on different criteria: Next, we address the question of how well criterion-specific individual models in the ensembles performed. Table 5 shows precision, recall, and F measure values for the Cascade Learner and the Ensemble-Boolean, which are top two performing ensembles, for each of the four criteria.

Note that the inputs to the ensemble components for Cascade Learner are filtered at each stage while all inputs are presented to the components of the Ensemble-Boolean. Performance generally reduces from Task 1 through 4 models for both architectures. Modeling a criterion is, therefore, more challenging as we go from criterion 1 through criterion 4. Specifically, the "positive" label prediction for Task 1 (format assessment) was 0.9876 F for the Cascade Learner, but the positive label performance of the Task 4 (rigor assessment) model reduced to 0.7780 F, a significant drop of nearly 21% in absolute terms. This performance reduction is even worse for Ensemble-Boolean which analyzes all samples. For the latter stage criteria, the Cascade Learner performs significantly better than the Ensemble-Boolean giving it an edge overall. Once again, we see how filtering out negative samples early improves the overall performance.

Table 5. Individual criterion-level performance for the top two ensembles using the partial (Del Fiol et al) dataset.

| Model | Task Number | Labels | P | R | F |
|---|---|---|---|---|---|
| Cascading Learner | Task 1- Format | Not O | 0.9433 | 0.8708 | 0.9056 |
| | | O | 0.9824 | 0.9928 | 0.9876 |
| | Task 2- HHC | F | 0.7463 | 0.8805 | 0.8078 |
| | | T | 0.9723 | 0.9334 | 0.9524 |
| | Task 3- Purpose | Not Tr | 0.9462 | 0.8679 | 0.9053 |
| | | Tr | 0.7489 | 0.8887 | 0.8128 |
| | Task 4- Rigor | F | 0.9723 | 0.8821 | 0.9250 |
| | | T | 0.6801 | 0.9089 | **0.7780** |
| Ensemble-Boolean | Task 1- Format | Not O | 0.9443 | 0.8679 | 0.9045 |
| | | O | 0.9821 | 0.9930 | 0.9875 |
| | Task 2- HHC | F | 0.6975 | 0.8852 | 0.7802 |
| | | T | 0.9728 | 0.9145 | 0.9427 |
| | Task 3- Purpose | Not Tr | 0.9487 | 0.8561 | 0.9000 |
| | | Tr | 0.7373 | 0.8972 | 0.8094 |
| | Task 4- Rigor | F | 0.9821 | 0.8260 | 0.8973 |
| | | T | 0.5894 | 0.9432 | **0.7254** |

*The impact of sampling ratio:* We show the impact of training sampling ratios in Table 6 for the ITL model. We see that when the positive samples are roughly 10 times oversampled and negatives are down sampled to the same size (i.e. 15K positive and 15K negative samples), ITL achieved its best F measure although the best recall and precision were achieved at other sampling ratios. The substantial performance improvement (almost 0.20 F measure in absolute terms) is an indication that the additional negative samples clearly helped ITL train better and is an indication of the larger research question, what is the optimal training ration and size when data is highly imbalanced.

Table 6. The impact of sampling ratios on ITL performance using the full (Marshall et al) dataset

| positive samples | negative samples | Precision | Recall | F Measure |
|---|---|---|---|---|
| 1587 | 1587 | 0.3825 | **0.9616** | 0.5472 |
| 15k | 15k | 0.6478 | 0.8601 | **0.7390** |
| 15k | 30k | 0.6848 | 0.7391 | 0.7109 |
| 15k | 49k | 0.7020 | 0.6591 | 0.6799 |
| 30k | 30k | 0.6959 | 0.7240 | 0.7097 |
| 30k | 49k | **0.7068** | 0.6471 | 0.6757 |
| 49k | 49k | 0.7051 | 0.6087 | 0.6534 |

*The Impact of sequence length and PT tag:* For the results reported so far, the sequence length was 256 and the PT tag was not used. We present the impact of these factors as an "ablation" study as shown in Table 7. To improve clarity, we discuss results for ITL and Cascade Learner (the top two performing models). We started with the base configuration for each model and then successively changed features. For ITL, we started with sequence length of 256 (which is 69$^{th}$ percentile) and equal number of positive and negative training samples without up sampling (1.5k and 1.5k) and, increased sequence length to 384 (92$^{nd}$ percentile), training samples to 15K-15K by up-sampling positives, and finally added PT tag text to the input. Similarly, for Cascade Learner, started with 256 sequence length and balanced training data at each component model, and increased sequence length to 384, and finally add PT tag text to the input.

Changing training samples by 10x, i.e. using 10x distinct negative samples, significantly improved performance of ITL through a big jump in precision and a small loss in recall, and a nearly 0.20 improvement in F measure in absolute terms. Subsequently, changing to sequence length of 384 worsened performance, but improved further after PT tag text was added to the input. We tried (not reported here) by adding PT tag without changing sequence length but it did not provide better performance than reported here with 384. We think the increased sequence length was not beneficial because of the noise from *null* padding but the PT tag text reduced the noise and also helped to improve precision.

As mentioned earlier, we always used balanced positive and negative samples at each component in Cascade Learner, and as we changed sequence length and added PT tag text to the input, the performance changes were similar to what we observed with ITL. Cascade Learner achieves the best overall performance with 384 sequence length and with the PT tag resulting in the highest overall precision (0.6686) and F measure (0.7639). ITL in the base configuration achieves the highest recall (0.9616) with lowest precision.

*Table 7. Ablation study: impact on performance as features are added/changed, conducted using the full dataset.*

| Model | Configuration | Precision | Recall | F Measure |
|---|---|---|---|---|
| ITL | 1.5K-1.5k, 256 | 0.3825 | **0.9616** | 0.5472 |
| | **15K-15K**, 256 | 0.6478 | 0.8601 | 0.7390 |
| | 15K-15K, **384** | 0.6396 | 0.8601 | 0.7337 |
| | 15K-15K, 384, **PT tag** | **0.6418** | 0.8796 | **0.7422** |
| Cascade Learner | Balanced, 256 | 0.6600 | 0.8796 | 0.7542 |
| | Balanced, **384** | 0.6436 | 0.8897 | 0.7469 |
| | Balanced, 384, **PT tag** | **0.6686** | **0.8910** | **0.7639** |

# Discussion

While ITL consistently achieved better recall (in multiple feature configurations) than the ensembles including Cascade Learner, the Cascade Learner achieved better precision than ITL and other architectures with relatively smaller loss of recall, consequently achieving the best F measure. Thus, modeling all constraints together seem to capture the diversity in relevant articles, but separately modeling the constraints and ensembling them in a cascade helps to improve the accuracy (entropy) of classification. This is analogous to query optimization in IR and in databases where rearranging the query terms is known to improve quality (in IR) and speed (in databases). This is an important finding of our study and generalizes beyond the particular task and dataset we used.

Among the individual constraint-specific models, the last one, i.e. deciding whether a study meets scientific "rigor" requirements, was seen to be the most challenging (due to its lowest F measure, see Table 4). Clearly, finding scientifically sound articles is not as easy as matching "Randomized Controlled Trial" or its simple equivalents in the text even with a state-of-the-art language model such as SciBERT. Therefore, one area of future work is to improve modeling of this semantically challenging task.

We compared our results with Del Fiol et al and Marshall et al, although certain experimental differences made it difficult to draw straightforward conclusions from the comparison. We used ITL or Cascade Leaner results for comparison depending on whether the comparison involved high recall or balanced precision-recall (i.e. highest F measure) results.

Del Fiol et al study reported only one set results for a neural network (CNN) model, which achieved 0.969 recall, 0.346 precision, and 0.510 F measure. Our ITL model achieved higher precision (0.509) at the same recall, and thus achieving higher F measure (0.6674) also. See Table 8. The study also reported results for a customized search filter, referred to as the McMaster's clinical query (CQ) balanced filter, that was meant to achieve the best balance between precision and recall. Compared to it our best ITL and Cascade models achieved distinctly higher F measures, i.e. 0.6700 and 0.7505 respectively versus 0.575 reported in the study. Lastly, compared to the McMaster's CQ broad query (meant to achieve highest recall) which achieved 0.224 precision at 0.984 recall, our ITL achieved 0.270 precision at a slightly higher 0.985 recall. While care should be taken in comparing specific details, it is clear that SciBERT based Cascade Learner achieved a distinctly better F measure, 0.7505 vs. 0.510, compared to the early neural network model, CNN. The performance difference is 0.2405 absolute F measure and a reduction of 49.1% in error rate relative to the CNN model performance.

*Table 8. Performance comparison with results reported in Del Fiol et al*

|  | Model | Parameters/Details | Performance | | |
|---|---|---|---|---|---|
|  |  |  | P | R | F |
| Comparison with CNN | CNN | Best result from Del Fiol et al | 0.3460 | 0.9690 | 0.5100 |
|  | Our ITL @ recall 0.969 | 1.5k-1.5k, 384, PT | 0.5090 | 0.9690 | 0.6674 |
| Comparison with balanced CQ filters | Customized text query | McMaster's clinical query balanced filter | 0.4090 | 0.9700 | 0.5750 |
|  | Our ITL Best | 1.5k-1.5k, 256, no PT | 0.5137 | 0.9627 | 0.6700 |
|  | Our Cascade Best | Balanced samples, 256, no PT | 0.6551 | 0.8783 | 0.7505 |
| High sensitivity CQ filters | Customized text query | McMaster's clinical query broad filter | 0.2240 | 0.9840 | 0.3650 |
|  | Our ITL @ recall 0.985 | 1.5k-1.5k, 384, PT | 0.2700 | 0.9850 | 0.4238 |

The Marshall et al study reported two sets of best precision and recall results: (1) Using SVM (support vector machines) in a voting ensemble with a PT tag classifier (see the study[9] for details) and recall fixed at 0.985; (2) Using CNN in a voting ensemble with the PT tag classifier and specificity fixed at 0.975. From these results, we calculated corresponding F measures. As shown in Table 9, our ITL model performed slightly better than the best performing high-sensitivity model (SVM + PT tag classifier) of the Marshall et al study. However, our ITL and Cascade Learner both outperformed the best F measure (i.e. the fixed specificity results) reported in the Marshall et al study. In particular, Cascade Learner achieved 0.7639 vs. 0.7058 F measure, an improvement of 0.0581 F measure in absolute terms which is 19.7% error rate reduction.

*Table 9. Marshall et al comparison*

|  | Model | Parameters/Details | Performance | | |
|---|---|---|---|---|---|
|  |  |  | P | R | F |
| Comparison of best model Performances | CNN + Voting with PT tag | Marshall et al best (which was at fixed specificity 0.975) | 0.5590 | 0.9570 | 0.7058 |
|  | Our ITL best | 15k-15k, 384, PT tag | 0.6418 | 0.8796 | 0.7422 |
|  | Our Cascade best | Balanced, 384, PT tag | 0.6686 | 0.8910 | **0.7639** |
| Comparison at fixed recall 0.985 | SVM + Voting with PT tag | Marshall et al best @ recall 0.985 | 0.2100 | 0.9850 | 0.3462 |
|  | Our ITL @ 0.985 recall | 1.5k-1.5k, 384, PT tag | 0.2180 | 0.9850 | **0.3570** |

# Conclusion

We made an extensive study of the latest neural network models for the task of filtering articles for systematic reviews and other meta-analysis using the Clinical Hedges dataset. The overall observation was that SciBERT based models offer superior performance in identifying scientifically sound articles compared to the early neural network models or feature-engineered models, even when the dataset is highly imbalanced (up to 1 to 32, positive to negative ratio).

Since filtering of such articles is often based on multiple criteria, we showed that a cascaded ensemble of criterion-specific models is superior in terms of achieving high precision-recall balanced (i.e. high F measure) results compared to standard ensembles of the criteria or a single model trained on all the criteria. This is a significant result that generalizes beyond this task and the dataset, and has a similarity to query optimization in IR and in databases.

The single integrated model however consistently achieved higher recall, suggesting that it would be a better architecture for high recall (sensitivity) results which may be acceptable at a low precision (i.e. at a high noise level). We also observed that changing the number and the ratio of negative samples could have a significant impact of the performance of the single integrated model. This raises yet again an important research question: what is the best number of negative samples and the ratio of positive and negative samples for optimum training of a model when the training data is highly imbalanced.

## Conflicts Interest: None